\def\year{2019}\relax
\documentclass[letterpaper]{article} 
\usepackage{aaai19}  
\usepackage{times}  
\usepackage{helvet}  
\usepackage{courier}  
\usepackage{url}  
\usepackage{graphicx,subcaption}  
\usepackage{float}
\usepackage{amsmath}
\DeclareMathOperator*{\argmin}{argmin} 
\usepackage{amssymb}
\usepackage{amsthm}
\usepackage{url}
\usepackage{multirow}
\usepackage{multicol}
\usepackage{xcolor}
\usepackage{diagbox}
\graphicspath{ {./manuscript/figures/} } 
 
\begin{document}
%
\title{Privacy-Preserving Image Classification in the Local Setting}

\author{Sen Wang,\textsuperscript{1}
J. Morris Chang\textsuperscript{1}\\
\textsuperscript{1}{University of South Florida}\\
senwang@mail.usf.edu, 
chang5@usf.edu}

\maketitle
\begin{abstract}
Image data has been greatly produced by individuals and commercial vendors in the daily life, and it has been used across various domains, like advertising, medical  and traffic analysis. Recently, image data also appears to be greatly important in social utility, like emergency response. However, the privacy concern becomes the biggest obstacle that prevents further exploration of image data, due to that the image could reveal sensitive information, like the personal identity and locations. The recent developed Local Differential Privacy (LDP) brings us a promising solution, which allows the data owners to randomly perturb their input to provide the plausible deniability of the data before releasing. In this paper, we consider a two-party image classification problem, in which data owners hold the image and the untrustworthy data user would like to fit a machine learning model with these images as input. To protect the image privacy, we propose to locally perturb the image  representation before revealing to the data user. Subsequently, we analyze how the perturbation satisfies $\epsilon$-LDP and affect the data utility regarding count-based and distance-based machine learning algorithm, and propose a supervised image feature extractor, DCAConv, which produces an image representation with scalable domain size. Our experiments show that DCAConv could maintain a high data utility while preserving the privacy regarding multiple image benchmark datasets.
\end{abstract}

\section{Introduction}

Image data has been greatly produced and shared in our daily life, and the usage of image data has been demonstrated through various applications, like preference analysis and advertising\cite{leon2012johnny}, medical \cite{xia2016privacy} and traffic\cite{chan2008privacy} analysis. Lately, image data becomes greatly important in social utility\cite{lepinski2016privacy} as well, like emergency response. For example, in the Boston Marathon bombing, the authorities spent lots of effort to gather onsite photos from people in that area, and pore through thousands of photos to search for the suspects. However, the privacy concern comes to be the biggest obstacle that prevents people sharing this information, since the image may reveal where they are and what they are doing. Thus, exploring the privacy preserving machine learning technique with regard to image data becomes paramount importance.

In this paper, we study an image classification problem, in which there are two parties\cite{kairouz2014extremal}, the image owners, who hold a set of images; the untrustworthy data user, who would like to fit a machine learning model regarding these images. As an example, in the Marathon bombing case, the authorities serves as the data user, who asks for the onsite images to search for the suspects, and the people that provided the images serve as the data owner, who wishes to support the investigation while preserving the privacy. During the last decade, Differential Privacy \cite{Dwork2006DP} is one of the most popular schemes for privacy protection, which prevents the inference about individuals from participation of computation. It defines a mechanism that the computation of the data is robust to any change of any individual sample by introducing uncertainty into the algorithm. Recently, Local Differential Privacy(LDP)\cite{duchi2013local}\cite{wang2017locally} extends the technique into a local setting, which provides a stronger privacy protection that allows data owners to perturb their input before releasing.

Differentially private machine learning in the local setting has been rarely studied. Recently, \cite{nguyen2016collecting} and \cite{pihur2018differentially} investigate the privacy-preserving distributed machine learning, in both work, the data owners interactively work with the data user to learn the model. Unlike these previous work, a ``non-interactive'' manner draws much of our interest that the data owners only perturb once and then share the data with the data user. Comparing to the interactive setting, the non-interactive way is more scalable and needs less communication cost. Randomized response\cite{warner1965randomized} is proposed to provide the plausible deniability for individuals responding to sensitive surveys. It has been proved that this mechanism satisfies $\epsilon$-LDP and is universally optimal for mutual information optimization problem\cite{kairouz2014extremal}, which provides a solid foundation for its usage in the area of machine learning. In this paper, we are interested in adopting randomized response to locally perturb the input and fit the count-based and distance-based machine learning models with such input. Specifically, we study the utility and privacy trade-off with respect to the Naive Bayes and KNN classifiers and show how the model utility is affected by the privacy parameters. Furthermore, in terms of image classification, we propose the DCAConv, a supervised image feature extractor, that improves the utility under the $\epsilon$-LDP.

Overall, the contributions of our work are three folds:

\begin{itemize}
\item We propose to use the LDP in the privacy preserving image classification problem. To the best of our knowledge, most LDP-based work focus on the frequency estimation of the categorical data, and we are the first to investigate the privacy preserving mechanism to fit the image data in the classification problem.
\item We analyze the utility and privacy trade-off regarding count-based and distance-based machine learning models, i.e. Naive Bayes and KNN classifiers. As a result, we show how the model utility is affected by the privacy parameters. 
\item We develop the DCAConv, a supervised image feature extractor, which represents the image with a scalable domain size. DCAConv is evaluated in terms of image classification through multiple benchmark datasets, like texture and facial datasets. The results confirm that DCAConv could maintain a high data utility while preserving the privacy.
\end{itemize}

The rest of paper is organized as follow. The preliminary is in Section II. The problem definition and proposed solution is introduced in Section III. We describe DCAConv, the supervised image feature extractor in Section IV. The experiment and evaluation is in Section V. The related work is presented in Section VI. Section VII presents the conclusion.
\section{Preliminary}
\subsection{Local Differential Privacy}
Differential privacy (DP) ensures that an adversary should not be able to reliably infer any individual sample from the computation of a dataset, even with unbounded computational power and access to every other samples. Given two data database $A,A'$, it is said that $A,A'$ are \textit{neighbors} if they differs on at most one row. The formal definition of a $(\epsilon,\delta)$-differential private mechanism over $A$ is defined below: 

\textit{Definition 1 }($\epsilon,\delta$)-differential privacy\cite{Dwork2006DP,dwork2014analyze}: A randomized mechanism $\mathcal{F}$ is $(\epsilon,\delta)$-differentially private if for every two neighboring database $A,A'$ and for any $\mathcal{O} \subseteq Range(\mathcal{F})$,
\begin{equation}
    Pr[\mathcal{F}(A)\in \mathcal{O}] \leq e^\epsilon Pr[\mathcal{F}(A')\in \mathcal{O}] + \delta
\end{equation}

where $Pr[\cdot]$ denotes the probability of an event, $Range(\mathcal{F})$ denotes the set of all possible outputs of the algorithm $\mathcal{F}$. The smaller $\epsilon,\delta$ are, the closer $Pr[\mathcal{F}(A)\in \mathcal{O}]$ and $Pr[\mathcal{F}(A')\in \mathcal{O}]$ are, and the stronger privacy protection gains. When $\delta=0$, the mechanism $\mathcal{F}$ is $\epsilon$-differentially private, which is a stronger privacy guarantee than $\epsilon,\delta$-differential privacy with $\delta>0$.

Similar to DP, a stronger setting of \textit{local privacy} is investigated by Duchi et al. \cite{duchi2013local}, namely Local Differential Privacy. It considers a scenario that a data curator needs to collect data from data owners, and infer the statistical information from these data. The data owner is assumed to trust no one, and would perturb the data before sharing with data curator. The formal definition is given below:

\textit{Definition 2 }$\epsilon$-Local Differential Privacy\cite{duchi2013local,wang2017locally}: A randomized mechanism $\mathcal{G}$ satisfies $\epsilon$-Local Differential Privacy ($\epsilon$-LDP) if for any \textit{input} $v_1$ and $v_2$ and for any $\mathcal{S} \subseteq Range(\mathcal{G})$:
\begin{equation}
    Pr[\mathcal{G}(v_1) \in \mathcal{S}] \leq e^\epsilon Pr[\mathcal{G}(v_2) \in \mathcal{S}]
\end{equation}

Comparing to DP, LDP provides a stronger privacy model, while entails more noise.

\subsection{Randomized Response}
Randomized Response\cite{warner1965randomized} is a decades-old technique which is designed to collect the statistical information regarding sensitive survey questions. The original technique is used for frequency estimation of a particular input. For social research, the interviewee is asked to provide an answer regarding the sensitive question, such as marijuana usage. To protect the privacy, the interviewee is asked to spin a spinner unobserved by the interviewer. Rather than answering the sensitive question directly, the interviewee only tells yes or no according to whether the spinner points to the true answer. Suppose that the interviewer wants to estimate the population of using marijuana and asks the interviewee if he has ever used the marijuana. And the probability that the spinner points to yes is $p$, then the spinner points to no is with probability $1-p$. Thus an unbiased maximum likelihood estimates of the true population proportion is given by $\frac{p-1}{2p-1}+\frac{n_1}{(2p-1)n}$, where $n$ is the number of participating interviewees and $n_1$ is the amount of reporting yes. It\cite{wang2016private} shows that this mechanism satisfies $(ln \frac{p}{1-p})$-LDP.

Randomized response could also be generalized to multiple choices, where the domain of the choices is $[d], d>2$. Suppose the randomized method is $\mathcal{G},\forall v,s \in [d]$, the perturbation is defined below:

\begin{equation}\label{equ:randmizedResponse}
Pr[\mathcal{G}(v)=s]= \begin{cases}
               p = \frac{e^\epsilon}{d-1+e^\epsilon}\text{, if } v=s\\
               q = \frac{1-p}{d-1} = \frac{1}{d-1+e^\epsilon}\text{, if } v \neq s
            \end{cases} 
\end{equation}

\begin{proof}
\cite{wang2017locally} For any inputs $v_1$ and $v_2$ and output $s$:
\begin{equation}
\frac{\mathcal{G}(v_1)=s}{\mathcal{G}(v_2)=s}\leq \frac{p}{q}=e^{\epsilon}
\end{equation}
\end{proof}
In terms of maintaining the original statistical information, even though \cite{kairouz2014extremal} proves that randomized response is universally optimal for mutual information and $f$-divergences optimization, there's no clear guide about how to employ this mechanism in particular machine learning algorithm.

\subsection{Discriminant Component Analysis}
Principal Component Analysis is a well known tool for unsupervised learning which finds the subspace that preserves the most variance of the original data, however, it is not effective in the supervised learning since the labeling information is ignored in the computation. Discriminant Component Analysis (DCA)\cite{kung2017discriminant}, basically a supervised PCA, is proposed recently as a complement for supervised learning, where the signal subspace components of DCA are associated with the discriminant power regarding the classification effectiveness while the noise subspace components of DCA are tightly coupled with the recoverability. Since the rank of the signal subspace is limited by the number of classes, DCA is capable to support classification using a small number of components. 

Consider a $K$-class data set consisting of $N$ samples $X=[x_{1}, x_{2}, \dots, x_{n}]^T$, in which $x_{i} \in {\Bbb R}^m$. Each sample $x_{i}$ associates with a class label $y_{i}$ where $y_{i}\in \{C_{1}, C_{2}, \dots, C_{k}\}$. Let $\mu$ denotes the centroid of the total mass, $\mu_k$ denotes the centroid of the samples in class $C_k$, and $N_k$ denotes the number of samples in class $C_k$. The signal matrix is represented by the between-class scatter matrix:
\begin{equation}
    S_{B}=\sum^{K}_{k=1}N_k(\mu_{k}-\mu)(\mu_{k}-\mu)^{T}
\end{equation}
The noise matrix is characterized by the following within-class scatter matrix:
\begin{equation}
    S_{W}=\sum^{K}_{k=1} \sum_{i:y_i = C_k} (x_i-\mu_k)(x_i-\mu_k)^{T}
\end{equation}
The total scatter matrix $\bar{S}$ can be written as follows:
\begin{equation}
    \bar{S}= S_{B}+S_{W}
\end{equation}
In DCA, two ridge parameters $\rho$ and $\rho'$ are incorporated in the noise matrix $S_W$ and the signal matrix $S_B$ as follows:
\begin{multicols}{2}
  \begin{equation}
    S_{B}'=S_{B}+\rho' \mathbb{I}
  \end{equation}\break
  \begin{equation}
    S_{W}'=S_{W}+\rho \mathbb{I}
  \end{equation}
\end{multicols}
where $\mathbb{I}$ is the identity matrix. Therefore, the regulated scatter matrix is denoted as:
\begin{equation}
    \bar{S}'= S_{B}'+S_{W}'=\bar{S}+(\rho+\rho') \mathbb{I}
\end{equation}
DCA performs spectral decomposition of the pre-whitened scatter matrix, $D_{DCA}$:
\begin{equation}
D_{DCA} = (S_{W}')^{-1}\bar{S}' = U\Lambda U^T
\end{equation}
where $\Lambda$ holds the the monotonically decreasing eigenvalues, and U holds the associated eigenvectors. To form a projection matrix, it chooses $k$ eigenvectors with the largest eigenvalues to form a projection matrix $W_{dca} \in {\Bbb R}^{m \times k}$, in which each column is an eigenvector. And the raw data is transformed by multiplying the projection matrix:
\begin{equation} 
    X_{proj}= X \times W_{dca}
\end{equation}
\section{Problem Definition and Solution Analysis}
\subsection{Problem Definition}
\begin{figure}
  \centering
    \includegraphics[width=0.5\textwidth]{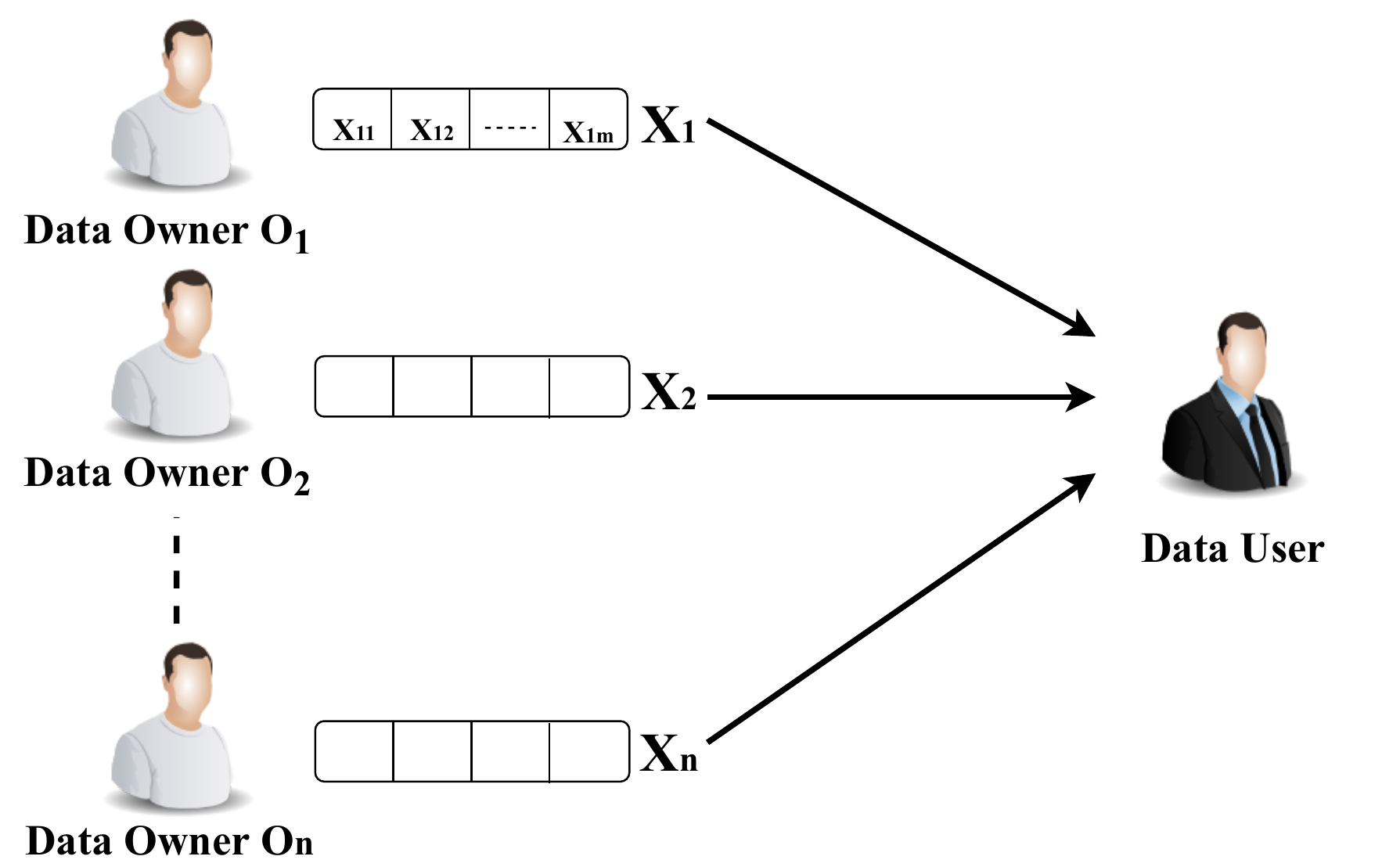}
  \caption{Problem Overview. An image held by Data Owner $O_i$ is represented as a vector $x_i = \{x_{i1},x_{i2},\dots,x_{im}\}, x_{ij} \in [d]$, each vector associates with a label $y_i$, the Data User would like to fit a classifier with such vectors from all $n$ Data Owners. To protect the privacy of $x_i$, a perturbation is required to satisfy $\epsilon_j$-LDP for the $j$th feature.}
  \label{fig:problem}
\end{figure}
As Fig.\ref{fig:problem} shows, in this paper, we consider the following problem: \textit{Given $n$ Data Owners, each Data Owner $O_i$ holds a set of images that are represented as a $m$-dimensional feature vectors. For ease of analysis, here it is assume that only one image is possessed by $O_i$ and it is denoted as $x_i = \{x_{i1},x_{i2},\dots,x_{im}\}, x_{ij} \in [d]$. Each image $x_i$  associates with a label $y_i$, for instance, the images held by each data owner is about the handwritten digit, and $y_i$ indicates the corresponding digit of $x_i$; the untrustworthy Data User would like to fit an image classifier with $x_i,y_i, i\in \{1,2,\dots,n\}$ as input while satisfying $\epsilon_j$-LDP for the $j$th feature of $x_i$}. It should be noted that the images held by each data owner are perturbed independently, so there are no difference to perturb a set of images or a single image at $O_i$. With respect to image classification, the effectiveness of both Naive Bayes and KNN classifiers have been demonstrated through previous work\cite{boiman2008defense,mccann2012local}, which are studied in this work.

For the threat model, the adversary is intended to learn individual image $x_i$, who could be either the untrustworthy data user, a participating data owner or an outside attacker. It assumes that adversaries might have arbitrary background knowledge and there might be a collusion among them. The goal is maintain the data utility in terms of image classification while protecting the data privacy.

\subsection{Frequency and Mean Estimation}

In this paper, we mainly investigate to use Randomized Response to satisfy $\epsilon_{ij}$-LDP for $x_{ij}$, which is perturbed independently. Given a $x_i,y_i$, we assume that the class label $y_{i}\in \{C_{1}, C_{2}, \dots, C_{k}\}$ and $x_i'$ is the vector after perturbation. Firstly, we show how to build the Naive Bayes classifier by leveraging the aggregation result of the perturbed input.

In our problem, for each training data $x_i$, $x_{ij}$ is independently perturbed by randomized response at $O_i$ and $x_i'$ is sent to the Data User. And the frequency of each appeared value $v$ is served as the building block of the classifier training, more specifically, the frequency of value $v$ in the $j$th feature that belongs to $C_k$ should be counted, in which is denoted as $c_j(v)$. However, due to the noise injected by randomized response, the observation doesn't reflect the true proportion of value $v$. Thus given the number of observation of $v$ in the $j$th feature, denoted as $\sum_jv$, $c_j(v)$ is estimated as\cite{wang2017locally}:

\begin{align}
\hat{c_j(v)} & = \frac{\sum_jv - nq_j}{p_j-q_j} \\
& = \frac{\sum_jv - n(1-p_j)/(d-1)}{p_j-(1-p_j)/(d-1)}
\end{align}

where $p_j=Pr[x_{ij}'=v]$, if $x_{ij}=v$, which refers to Equ.\ref{equ:randmizedResponse}, and $n$ is the number of observations that in $C_k$. It can be further shown that $\hat{c_j(v)}$ is an unbiased estimator. Assuming $f_{v,j}$ is the true proportion that value $v$ occurs in $j$th feature, $E[\hat{c_j(v)}]$ would be:

\begin{align}
E[\hat{c_j(v)}] & = E[\frac{\sum_jv - nq_j}{p_j-q_j}] \\
& = \frac{nf_{v,j}p_j+n(1-f_{v,j})q_j-nq_j}{p_j-q_j} \\ 
& = nf_{v,j}
\end{align}

And $Var[\hat{c_j(v)}]$ is given as below:
\begin{align}
Var[\hat{c_j(v)}] & = \frac{nq_j(1-q_j)}{(p_j-q_j)^2} + \frac{nf_{v,j} (1-p_j-q_j)}{p_j-q_j} \\
& = n(\frac{d-2+e^{\epsilon_j}}{(e^{\epsilon_j}-1)^2} + \frac{f_{v,j}(d-2)}{e^{\epsilon_j}-1})
\end{align}

where $Var[\hat{c_j(v)}]$ is in $O(\frac{nd}{e^\epsilon})$. Furthermore, $\hat{\mu_j}$ is the estimated mean of feature $j$ given the frequency estimation of each appeared value:
\begin{equation}
\hat{\mu_j} = \frac{\sum_vv\hat{c_j(v)}}{n}
\end{equation}
Correspondingly, $E[\hat{\mu_j}]$ and $Var[\hat{\mu_j}]$ are computed as follow:

\begin{align}
E[\hat{\mu_j}] & = E[\frac{\sum_vv\hat{c_j(v)}}{n}] \\
& =  \frac{E[\sum_vv\hat{c_j(v)]}}{n}\\  
& = \frac{\sum_vvE[\hat{c_j(v)]}}{n} \\
& = \sum_vvf_{v,j}
\end{align}

\begin{align}
Var[\hat{\mu_j}] & = Var[\frac{\sum_vv\hat{c_j(v)}}{n}] \\
& = \frac{Var[\sum_vv\hat{c_j(v)]}}{n^2}\\  
& = \frac{\sum_vv^2Var[\hat{c_j(v)]}}{n^2}
\end{align}

It can be seen that $\hat{\mu_j}$ is an unbiased estimator and $Var[\hat{\mu_j}]$ is in $O(\frac{d}{ne^\epsilon})$. And the Mean-Squared-Error (MSE) of $\hat{\mu_j}$ is equal to $Var[\hat{\mu_j}]$. In practical, the unbiased estimator for the sum of counts is adopted, which is $\sum_v\hat{c_j(v)}$ and it results in the same order of MSE.

A different mean estimator is given by replacing the denominator $n$ with the corrected sum of frequencies, where $\hat{\mu_j^{rr}}$:

\begin{equation}
\hat{\mu_j^{rr}} = \frac{\sum_vv\hat{c_j(v)}}{\sum_v\hat{c_j(v)}}
\end{equation}
Correspondingly, by using multivariate Taylor expansions, $E[\hat{\mu_j^{rr}}]$ and $Var[\hat{\mu_j^{rr}}]$ are computed as follow:

\begin{align}
& E[X]=E[\sum_vv\hat{c_j(v)}] = \sum_vE[v\hat{c_j(v)}] = \sum_vvE[\hat{c_j(v)}] \leq nd^2 \\
& var[X]=var[\sum_vv\hat{c_j(v)}] = \sum_vvar[v\hat{c_j(v)}] = \sum_vv^2var[\hat{c_j(v)}] \\
& E[Y]=E[\sum_v\hat{c_j(v)}] = \sum_vE[\hat{c_j(v)}] = n \\
& var[Y]=var[\sum_v\hat{c_j(v)}] = \sum_vvar[\hat{c_j(v)}] = nd(\frac{d-2+e^{\epsilon_j}}{(e^{\epsilon_j}-1)^2})
\end{align}

\begin{align}
& E[\frac{X}{Y}] \approx \frac{E[X]}{E[Y]}-\frac{cov[X,Y]}{E[Y]^2}+\frac{E[X]}{E[Y]^3}var[Y] \\
& \approx \frac{E[X]E[Y]^2}{E[Y]^3}-\frac{cov[X,Y]E[Y]}{E[Y]^3}+\frac{E[X]var[Y]}{E[Y]^3} \\
& \approx \frac{E[X]E[Y]^2-cov[X,Y]E[Y]+E[X]var[Y]}{E[Y]^3} \\
& \leq \frac{n^3d^2-ncov[X,Y]+n^2d^3\frac{d-2+e^{\epsilon_j}}{(e^{\epsilon_j}-1)^2}}{n^3} \\
& var[\frac{X}{Y}] \approx \frac{var[X]}{E[Y]^2}-\frac{2E[X]}{E[Y]^3}cov[X,Y]+\frac{E[X]^2}{E[Y]^4}var[Y] \\
& \approx \frac{var[X]E[Y]^2}{E[Y]^4}-\frac{2E[X]E[Y]cov[X,Y]}{E[Y]^4}+\frac{E[X]^2var[Y]}{E[Y]^4}\\
& \approx \frac{var[X]E[Y]^2-2E[X]E[Y]cov[X,Y]+E[X]^2var[Y]}{E[Y]^4} \\
& cov[X,Y] = E[XY] - E[X]E[Y]\\
& MSE[\frac{X}{Y}] = Var[\frac{X}{Y}] + (E[\frac{X}{Y}]-\frac{X}{Y})^2
\end{align}

\subsection{Nearest Centroid-Based Classifier}
The Nearest Centroid Classifier (NCC) is a classification model that assigns to testing data the label of the class of training data whose mean is closest to the observation in terms of certain distance metrics. Given training data $x_i,y_i, i\in \{1,2,\dots,n\}$, the per-class centroid is computed as follows:
\begin{equation}
\mu_k = \frac{1}{N_k}\sum_{i:y_i=C_k}x_i
\end{equation}
where $N_k$ is the number of training data that belonging to class $C_k$. For a testing data $x_t$, the predicted class is given by,
\begin{equation}
\hat{y_t} = \argmin_k\ell_p(x_t,\mu_k)
\end{equation}
where $\ell(\cdot)$ is the $p$-norm distance between $x_t$ and $\mu_k$. To build the NCC from the perturbed data, the estimated centroid is adopted:
\begin{equation}
\hat{\mu_k} = \frac{\sum_{v\in C_k}v\hat{c_j(v)}}{N_k}
\end{equation}

\section{DCAConv - Supervised Image Feature Extraction}

\begin{figure*}[ht!]
  \centering
    \includegraphics[width=0.95\textwidth]{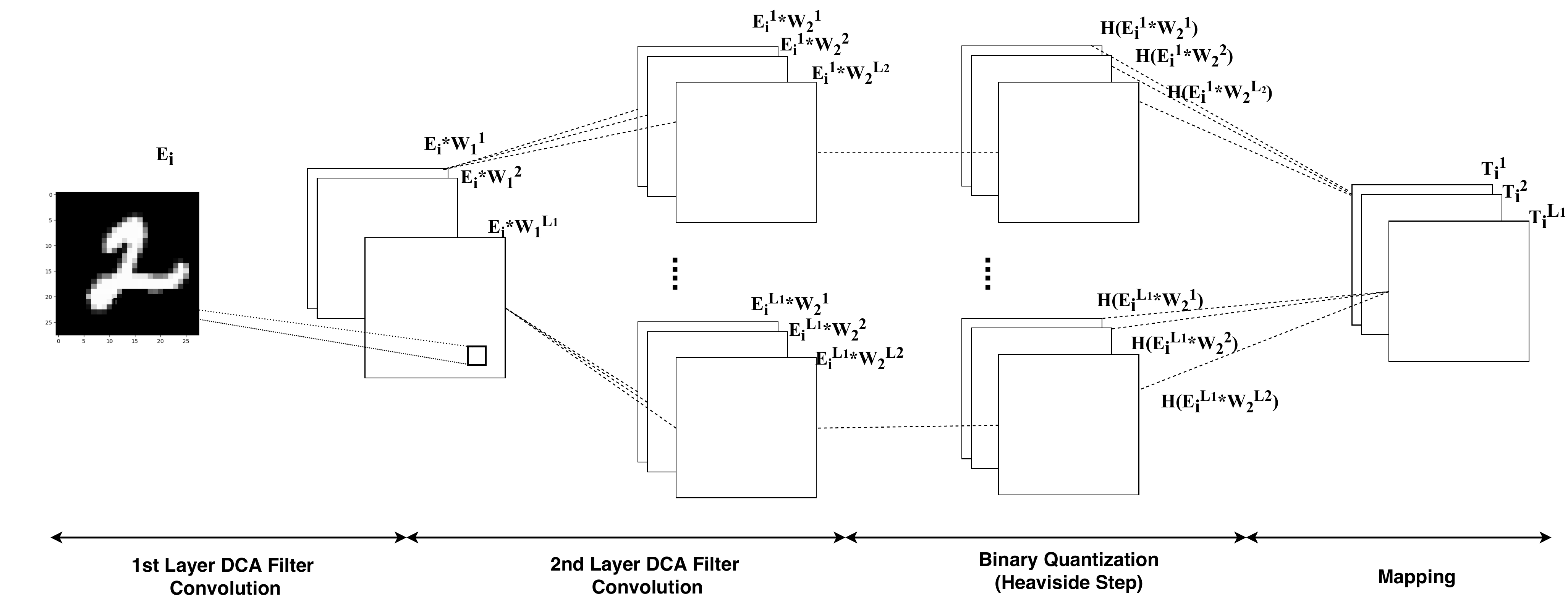}
  \caption{DCAConv: Supervised Image Feature Extractor}
  \label{fig:extractor}
\end{figure*}
As analyzed in previous section, for both Naive Bayes and KNN classifiers, we show that the domain size of the input determines the utility gain regarding the perturbation of randomized response. For Naive Bayes, the utility can be measured by the variance of the estimated frequency. For KNN classifier, the utility is measured by the probability of preserving the true proximity between the testing data points and the perturbed training points. In terms of randomized response, it is readily to see that the utility gain is maximized when domain size $d=2$. However, the binary format limits the transited information, especially for the image data. On the other hand, increasing $d$ asks for more price to pay with respect to privacy. Thus the domain size $d$ is the very key factor for the trade-off between the utility and privacy.

For image data, the naive way to protect the privacy is to apply randomized response to the pixel feature, nevertheless the utility is severely damaged due to the large domain size. Thus the challenge is to find a proper representation of the image data. On one hand, the representation should preserve as much information as possible; on the other hand, the domain size should provide a balance between the utility and privacy. The success of CNN\cite{krizhevsky2012imagenet} on image classification draws much attention in the last decade, which transforms the image into multiple level representations by extracting the abstract semantics. However, it is hard to apply randomized response to the CNN directly, the reason is that CNN usually needs continuous number in the data flow. Furthermore, it also relies on either Softmax or Sigmoid layer to determine the final output distribution, where the model is trained to optimize for specific loss functions. Inspired by PCANet\cite{chan2015pcanet}, we are interested in designing a mechanism to transform the image to a representation that has a scalable domain size. In the meanwhile, the domain size could be as small as possible to narrow the impact of randomized response in terms of utility. The outcome is DCAConv, as Fig.~\ref{fig:extractor} shows, which consists of the convolution layer, the binary quantization layer and the pooling layer. The convolution filter is computed using DCA and the generated discriminant components serve as the filter. The quantization layer converts the output of convolution operation to binary bits. Finally, images are represented with the mapping of the binary bits and passed through the pooling layer. The details are presented below:

\subsection{DCA Convolution Layer}
Suppose there are $N$ images $\{E_i\}^N_{i=1}$, each image has the same size $h \times w$, the patch size is $k_h \times k_w$, the number of filters in the $r$th convolution layer is $L_r$. $E_i$ is denoted as $[e_{i1}, e_{i2}, \dots , e_{i\widetilde{h}\widetilde{w}}]$, and $e_{ij}\in {\Bbb R}^{k_h\times k_w}$ denotes the $j$th vectorized patch in $E_i$, where $\widetilde{h} = (h-k_h+2z)/t+1, \widetilde{w} = (w-k_w+2z)/t+1$, and $t$ is the size of stride, $z$ is the the size of zero padding. Additionally, assuming there are $K$ classes, each image $E_i$ associates with a class label $y_i$ where $y_i\in \{C_{1}, C_{2}, \dots, C_{k}\}$, and each class $C_k$ contains $N_k$ images. For the first convolution layer, the overall patch mean $\mu$ and within class patch mean $\mu_k$ is calculated as below:
\begin{align}
  \mu & = \frac{\sum_{i=1}^N\sum_{j=1}^{\widetilde{h}\widetilde{w}} e_{ij}}{N} \\ 
  \mu_k & = \frac{\sum_{i\in C_k}^N\sum_{j=1}^{\widetilde{h}\widetilde{w}} e_{ij}}{N_k}
\end{align}
Then the between-class scatter matrix $S_B$, the within class scatter matrix $S_W$ and regularized scatter matrix are calculated as below:
\begin{align}
  S_B &= \sum^{K}_{k=1}N_k(\mu_k-\mu)(\mu_k-\mu)^T \\ 
  S_W &= \sum^{K}_{k=1} \sum_{i \in C_k} (E_i-\mu_k)(E_i-\mu_k)^T \\
  \bar{S}' &= S_{B}'+S_{W}'=S_{B}+\rho' \mathbb{I} + S_{W}+\rho \mathbb{I}
\end{align}
Once getting $S_W$ and $\bar{S}'$, DCA is performed over the patches and compute a set of discriminant components as the orthonormal filters, where:
\begin{equation}
(S_{W}')^{-1}\bar{S}' = U\Lambda U^T
\end{equation}
where $U$ holds the associated eigenvectors. The leading eigenvectors capture the main discriminant power of the mean-removed training patches. Then the first $L_1$ eigenvectors are taken as the orthonormal filters, where the $l$th filter is represented as $W^l_1\in {\Bbb R}^{k_h\times k_w}$, in the first convolution layer. 

Similar to CNN, multiple convolution layers could be stacked following the first layer, as Fig.~\ref{fig:extractor} shows. Given a second convolution layer as example, it repeats the similar procedure as in the first one. Given the $l_1$th filter in the first layer, for image $E_i$:
\begin{equation}
E_i^{l_1} = E_i * W^{l_1}_1, i = 1,2,\dots,N, l_1 = 1,2,\dots,L_1
\end{equation}

where $*$ denotes the 2D convolution. It could use zero padding on $E_i^{l_1}$ to make it has the same size as $E_i$. After that, the patch mean of $E_i^{l_1}$ is removed, then it performs DCA over the concatenated patches and computes the filters for the second convolution layer. Once DCA is finished, the first $L_2$ eigenvectors are taken as the filters, the $l$th filter in the second layer is represented as $W^l_2,W^l_2\in {\Bbb R}^{k_h\times k_w}$. For each $E_i^{l_1}$, there are $L_2$ filters to be convolved, as Fig.~\ref{fig:extractor} shows. 

\subsection{Binary Quantization}
After second convolution layer, there are $L_1\times L_2$ output in total, and they are converted to binary representation using the Heaviside step function $H(\cdot)$:
\begin{equation}
H(x) = 
\begin{cases}
    0, x \leq 0 \\
    1, x > 0   
\end{cases}
\end{equation}
where $x=E_i^{l_1}*W^{l_2}_2,l_1 = 1,2,\dots,L_1,l_2 = 1,2,\dots,L_2$.

\subsection{Mapping}
Once the output of the second convolution layer is converted to binary, for $E_i^{l_1}$, it obtains $\{H(E_i^{l_1}*W^{l_2}_2)\}_{l_2=1}^{L_2}$, which is a vector of $L_2$ binary bits. Since $E_i^{l_1}$ has the same size as $E_i$ with zero padding, for each pixel, the vector of $L_2$ binary bits is viewed as a decimal number of $L_2$ digits. It ``maps'' the output back into a single integer:
\begin{equation}
T_i^{l_1} = \sum^{L_2}_{l_2=1}2^{l_2-1}H(E_i^{l_1}*W^{l_2}_2)
\end{equation}

\subsection{Max Pooling}
The output of the binary quantization layer are $L_1$ ``images'' with each ``pixel'' of $L_2$ digits. To further reduce the size of the representation, for $T_i^{l_1}$, a max pooling layer is applied after the binary quantization layer.

\subsection{Perturbation}
The randomized response is applied to the output of the pooling layer, and the domain size $d$ is determined by $L_2$, where $d=2^{L_2}$. As we mentioned early, in DCA, the rank of the signal space is limited by the number of classes. More specially, $L_r \leq K$, where $L_r$ is the number of filters in $r$th convolution layer and $K$ is the number of classes. By employing DCA in the design, the domain of the final output is scalable, moreover, DCA also guarantees a high utility with a small number of components.
\section{Experiment}
We evaluate DCAConv over three popular benchmark image datasets, the MNIST, Fashion-MNIST\footnote{https://github.com/zalandoresearch/fashion-mnist}, YaleB\footnote{vision.ucsd.edu/~iskwak/ExtYaleDatabase/ExtYaleB.html}.

\textbf{The MNIST dataset} contains $28 \times 28$ grayscale images of handwritten digits that from 0 to 9, in which there are 60,000 samples for training and 10,000 samples for testing.

\textbf{The Fashion-MNIST dataset} contains the grayscale article images that each sample associates with a label from 10 classes. The dataset is intended to replace the overused MNIST dataset, and shares the same image size and structure of training and testing splits.

\textbf{The YaleB face dataset} contains grayscale photos of 27 subjects and each one has 594 in average, we select samples based on the light condition and the head position, which results in around 250 samples per subject. The face is extracted and scaled to 70 $\times$ 70.

To study the utility and privacy trade-off in terms of the domain size $d$ and privacy parameter $\epsilon$, we apply randomized response to multiple image representations, like raw pixel and  Histogram of Oriented Gradients (HOG), the perturbed input is then used to fit the Naive Bayes and KNN classifiers. For DCAConv, the domain size is altered by choosing different number of filters in the last convolution layer and the classification result demonstrates the effectiveness of the DCAConv over other image feature representations. All experiments are repeated 10 times, the mean and standard deviation of the classification accuracy are drawn in figures.
\begin{figure*}[ht!]
\centering
\begin{subfigure}[t]{1\textwidth}
\begin{subfigure}[t]{.32\textwidth}
\centering
\includegraphics[width=1\textwidth]{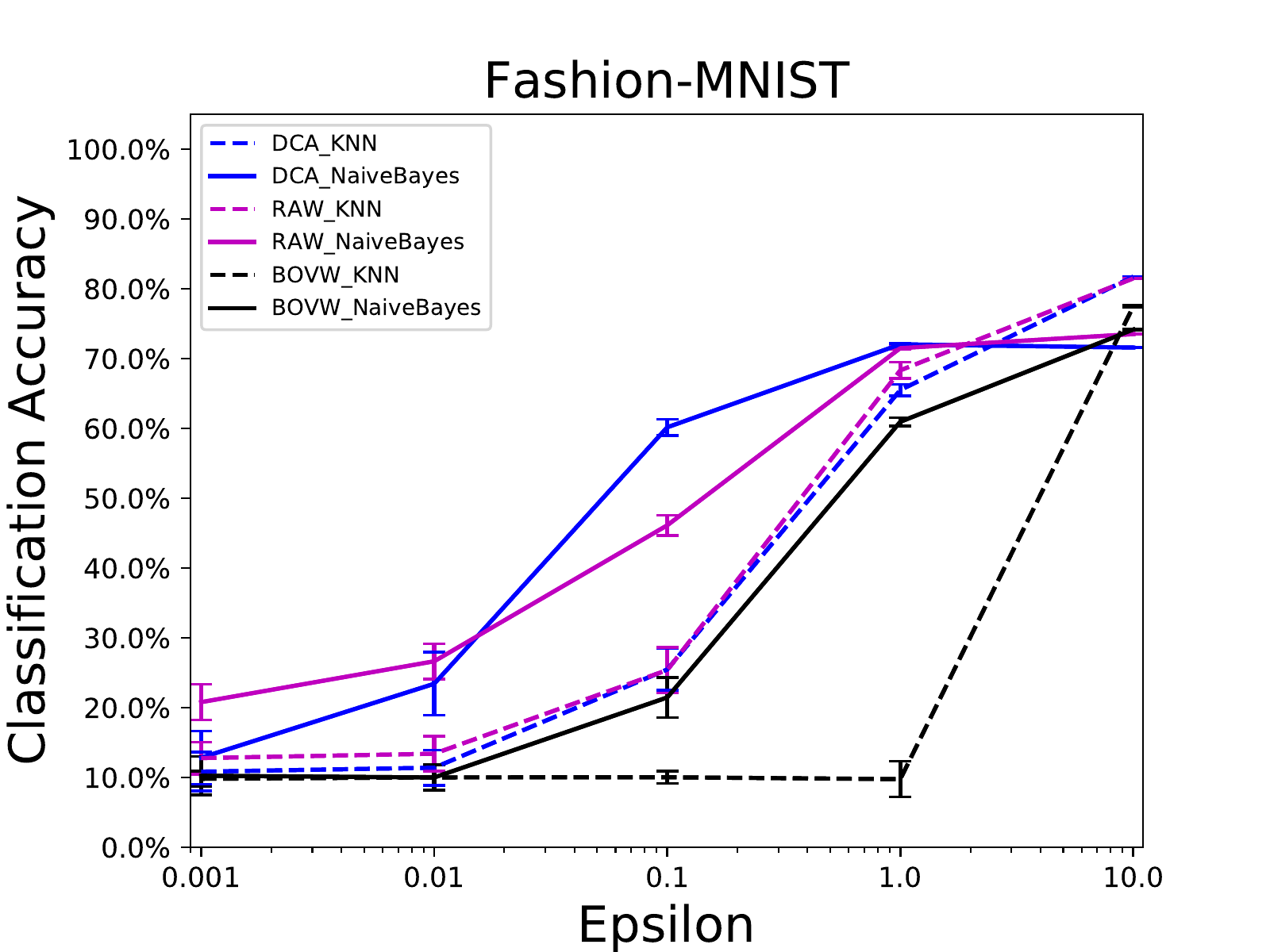}
\caption{}
\end{subfigure}
\begin{subfigure}[t]{.32\textwidth}
\centering
\includegraphics[width=1\textwidth]{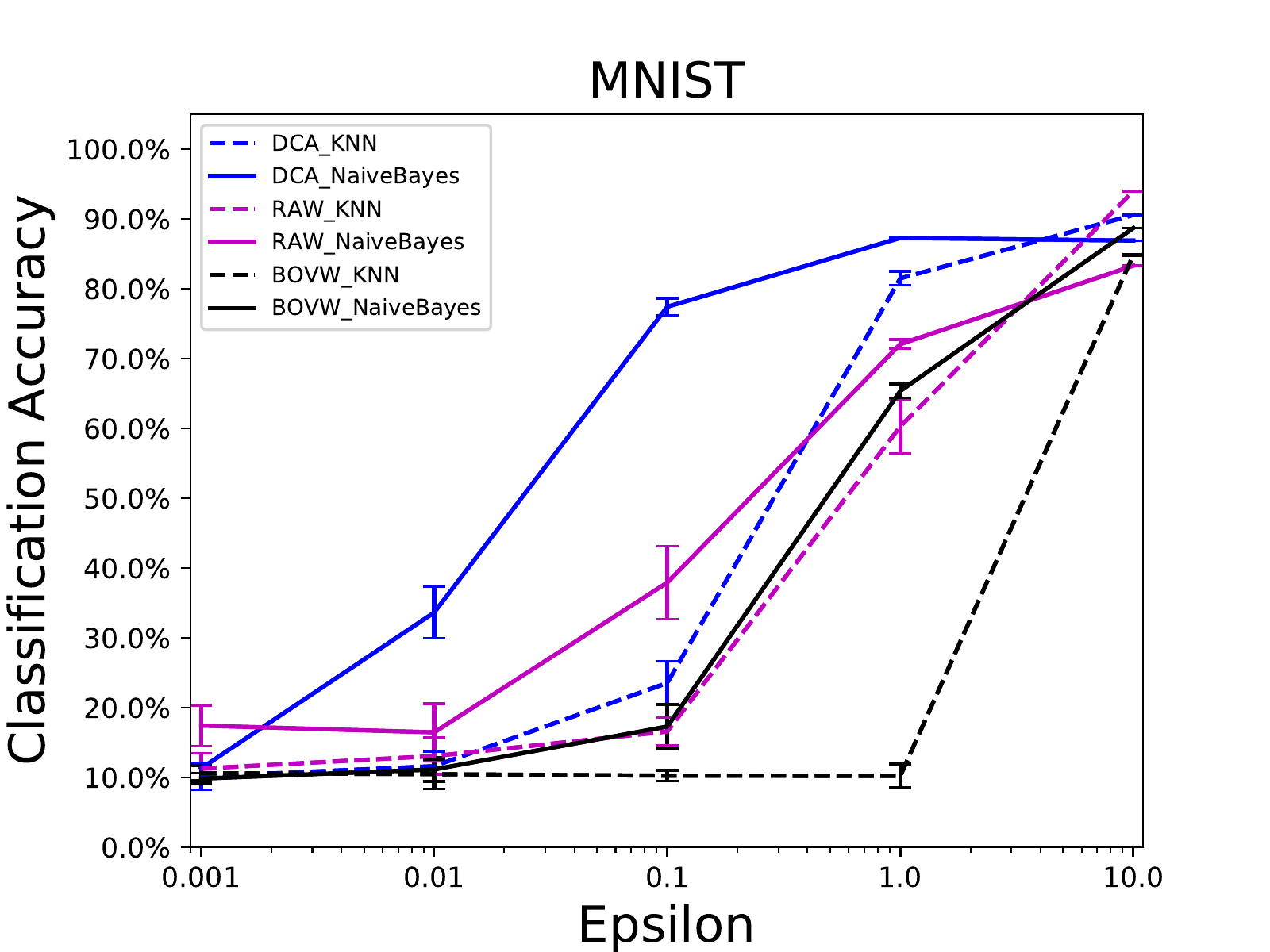}
\caption{}
\end{subfigure}
\begin{subfigure}[t]{.32\textwidth}
\centering
\includegraphics[width=1\textwidth]{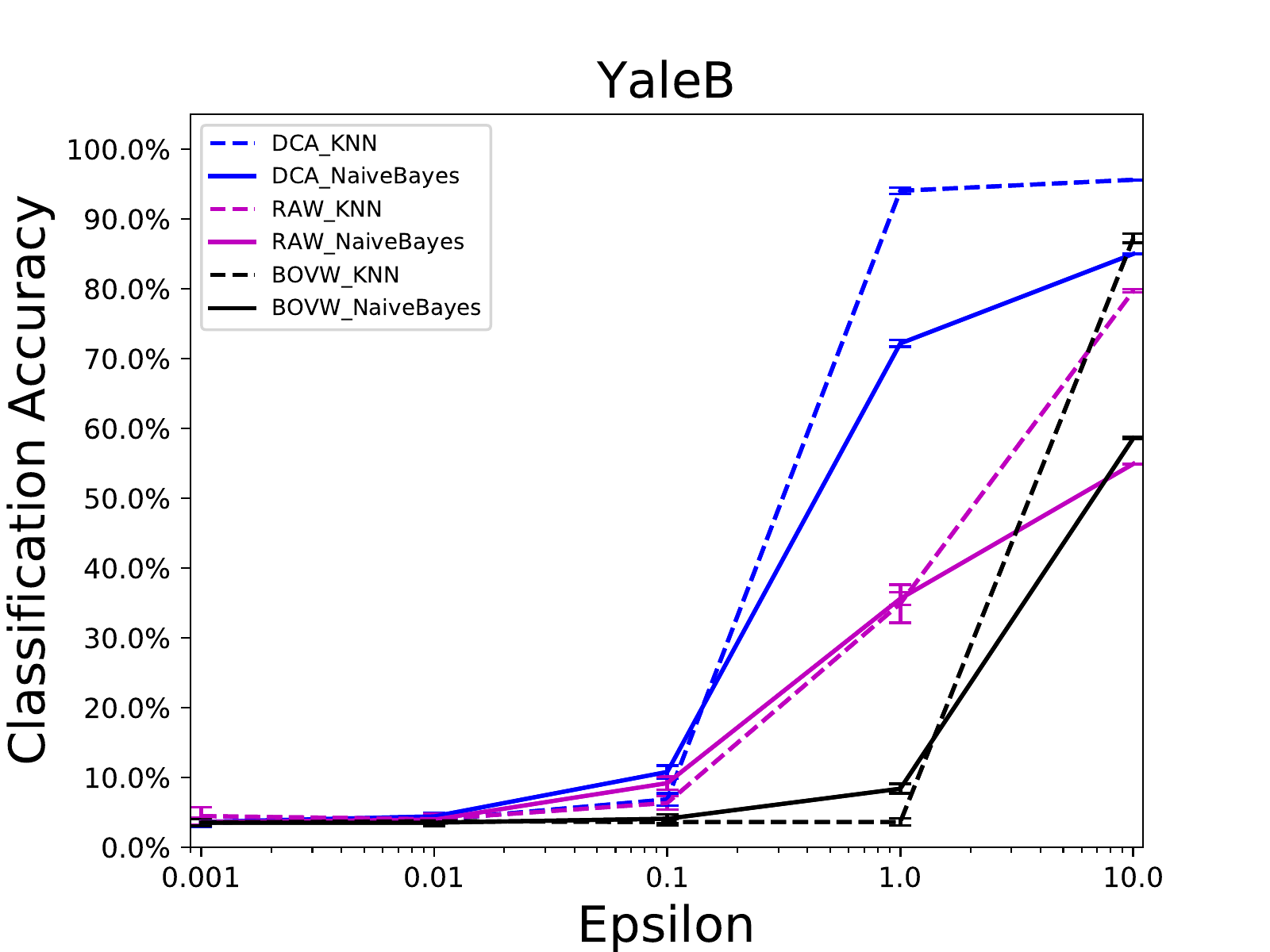}
\caption{}
\end{subfigure}
\end{subfigure}
\medskip
\caption{Comparison of DCAConv, pixel feature and BOVW, DCA\_KNN represents the classification result of DCAConv in KNN, RAW\_KNN represents the classification result of pixel feature in KNN, BOVW\_KNN represents the classification result of the BOVW in KNN. For DCAConv and pixel feature, the output domain $d$=16. For BOVW, MNIST has $d=40$, Fashion-MNIST has $d=39$ and YaleB has $d=44$. DCAConv hyper-parameters, filter size: 7$\times$7, filter stride size: 1$\times$1, $L_1$:5, $L_2$:4, pooling window size: 2$\times$2, pooling stride size: 1$\times$1.}
\label{fig:dca_vs_raw}
\end{figure*}
\subsection{DCAConv vs Pixel vs HOG}

We first measure the classification accuracy with three image representations, the raw pixel, DCAConv and HOG, where a Bag of Visual Words (BOVW) model is associated with the HOG descriptor. More specifically, HOG descriptor provides a fixed length of the image representation, while each feature is contrast-normalized to improve the accuracy. Once the descriptor is generated, a K-Means clustering is performed over the whole descriptors and the center of each cluster is used as the visual dictionary’s vocabularies. Finally, the frequency histogram is made from the visual vocabularies and is served as BOVW. Unlike DCAConv, the domain size of the BOVW is fixed once the number of clusters is determined and there is no significant difference observed after tuning the number of clusters. In the experiment, the domain size of the BOVW is determined by the maximum frequency observed among all features, e.g. MNIST has a domain size of 40, Fashion-MNIST has a size of 39 and YaleB has a size of 44, otherwise, the probability of answering honestly would be uneven among all features. For DCACov, there are two convolution layers and various number of $L_2$ filters is tested. To make a fair comparison, the raw pixel is downscaled to have the same size as the DCAConv output domain. For example, the raw pixel value is downscaled to [0,15] using a MinMaxScaler\footnote{http://scikit-learn.org/stable/modules/generated/\\sklearn.preprocessing.MinMaxScaler.html} to match the DCAConv with 4 $L_2$ filters. For the information loss caused by the downscaling, we don't observe a significant degrading in terms of the classification accuracy. Fig.~\ref{fig:dca_vs_raw} displays the accuracy regarding Naive Bayes and KNN classifiers, the horizontal specifies the $\epsilon$ for each individual feature, and the vertical axis shows the classification accuracy. 

From the figure, it could be seen that, both classifiers perform slightly better than random guessing when $\epsilon \leq 0.01$, and the accuracy increases as $\epsilon\geq 0.1$. When $\epsilon$ grows to $1.0$, for DCAConv and raw pixel, both classifiers provide reasonably good accuracy, however, the BOVW still suffers from a low accuracy as the domain size is much larger than the other two representations. Tab.~\ref{tab:dcaconv_d_16} provides the KNN accuracy of DCAConv when $\epsilon \geq 0.1$, where the last column gives the ground truth as no randomized response applied. As analyzed in Section III, the probability of being the true $\ell_2$ distance is bigger than $\frac{1}{2}$ as $\epsilon \geq ln(d+1)$, and it confirms that the accuracy approximates to the ground truth given that $\epsilon$, e.g. for all 3 datasets, the accuracy is within $5\%$ of the ground truth as $\epsilon \geq 2.83$. The comparison between Naive Bayes and KNN classifiers shows that Naive Bayes achieves higher accuracy when $\epsilon \leq 1.0$, where it may due to that the noise is eliminated by the frequency estimator, but the KNN still suffers to a low probability to preserve the true proximity of the data points given the same $\epsilon$. However, as $\epsilon$ increases, the perturbation becomes less, and the KNN starts to dominate classification accuracy.

Finally, for the texture dataset, the downscaled pixel feature provides a compatible accuracy with DCAConv. However, for the more complicated data, like facial data, the pixel fails to provide a good discriminant capacity in the KNN algorithm; under the same $\epsilon$, BOVW performs worse than the other two representations due to the large domain size.

\begin{table*}[t]
\centering
\begin{tabular}{|c|c|c|c|c|c|c|c|c|c|c|c|}
\hline
\multicolumn{1}{|l|}{} & \diagbox{dataset}{$\epsilon$}      & 0.1            & 0.5            & 1.0            & 1.5            & 2.0            & 2.5            & \textbf{3.0}            & 3.5            & 4.0            & $\infty$ \\ \hline
\multirow{3}{*}{Naive Bayes} & MNIST         & 77.52 & 86.35 & 87.15 & 87.19 & 87.07 & 86.97 & 86.92 & 86.92 & 86.90 & 86.90 \\ \cline{2-12} 
                             & Fashion-MNIST & 58.96 & 68.27 & 68.71 & 68.88 & 68.89 & 68.94 & 68.94 & 68.90 & 68.90 & 68.80 \\ \cline{2-12} 
                             & YaleB &  10.52 &   44.27    &  68.29     &  75.32     & 78.89       &  79.43     &  80.63     &  80.70    &   81.32    &  81.71  \\ \hline
\multirow{3}{*}{KNN}   & MNIST  & 24.72  & 68.92 & 81.96 & 86.05 & 88.00 & 89.37 & \textbf{89.95} &90.27 & 90.46 &  90.50 \\ \cline{2-12} 
                       & Fashion-MNIST  & 20.56 & 48.58 & 57.35 & 63.40 & 68.21 & 71.54 & \textbf{74.14} & 75.64 & 76.73 & 78.70 \\ \cline{2-12} 
                       & YaleB & 6.01  & 49.56  & 90.76 & 95.18 & 95.88 & 95.91 & \textbf{95.88} & 95.77 & 95.77 & 95.92 \\ \hline
\end{tabular}
\caption{Classification accuracy ($\%$) with DCAConv, where $d=16$ and $ln(d+1) \approx 2.83$. The last column provides the ground truth as no randomized response applied. Due to the space limit, the standard deviation of the measured accuracy is omitted here.}
\label{tab:dcaconv_d_16}
\end{table*}

\begin{table*}[t]
\centering
\begin{tabular}{|c|c|c|c|c|c|c|c|c|c|c|c|}
\hline
\multicolumn{1}{|l|}{} & \diagbox{dataset}{$\epsilon$}      & 0.1            & 0.5            & 1.0            & \textbf{1.5}            & 2.0            & 2.5            & 3.0            & 3.5            & 4.0            & $\infty$\\ \hline
\multirow{3}{*}{Naive Bayes} & MNIST         & 76.38 & 77.30 & 77.21 & 77.10 & 77.04 & 76.96 & 76.91 & 76.89 & 76.83 & 76.60 \\ \cline{2-12} 
                             & Fashion-MNIST & 58.96 & 68.27 & 68.71 & 68.88 & 68.89 & 68.94 & 68.94 & 68.90 & 68.90 & 68.80 \\ \cline{2-12} 
                             & YaleB         &  24.01     &   44.76    &  43.94     &  43.38     & 43.16       &   42.91    &   42.86    &  42.74     &  42.74     & 42.74      \\ \hline
\multirow{3}{*}{KNN}   & MNIST         &  65.96               &    85.25            &   88.27             &    \textbf{89.21}            &   89.69             &    89.95            &    90.04            &    90.19            & 90.24                &  90.20  \\ \cline{2-12} 
                       & Fashion-MNIST         & 69.66  & 70.90 & 70.62 & \textbf{70.52} & 70.43 & 70.39 & 70.39 & 70.37 & 70.32 & 70.30 \\ \cline{2-12} 
                       & YaleB & 79.33 & 95.50 & 95.35 & \textbf{95.46} & 95.58 & 95.55 & 95.61 &    95.66 & 95.75 & 95.80 \\ \hline
\end{tabular}
\caption{Classification accuracy ($\%$) with DCAConv, where $d=2$ and $ln(d+1) \approx 1.10$. The last column provides the ground truth as no randomized response applied. Due to the space limit, the standard deviation of the measured accuracy is omitted here.}
\label{tab:dcaconv_d_2}
\end{table*}

\subsection{DCAConv vs BNN}
\begin{figure*}[ht!]
\centering
\begin{subfigure}[t]{1\textwidth}
\begin{subfigure}[t]{.32\textwidth}
\centering
\includegraphics[width=1\textwidth]{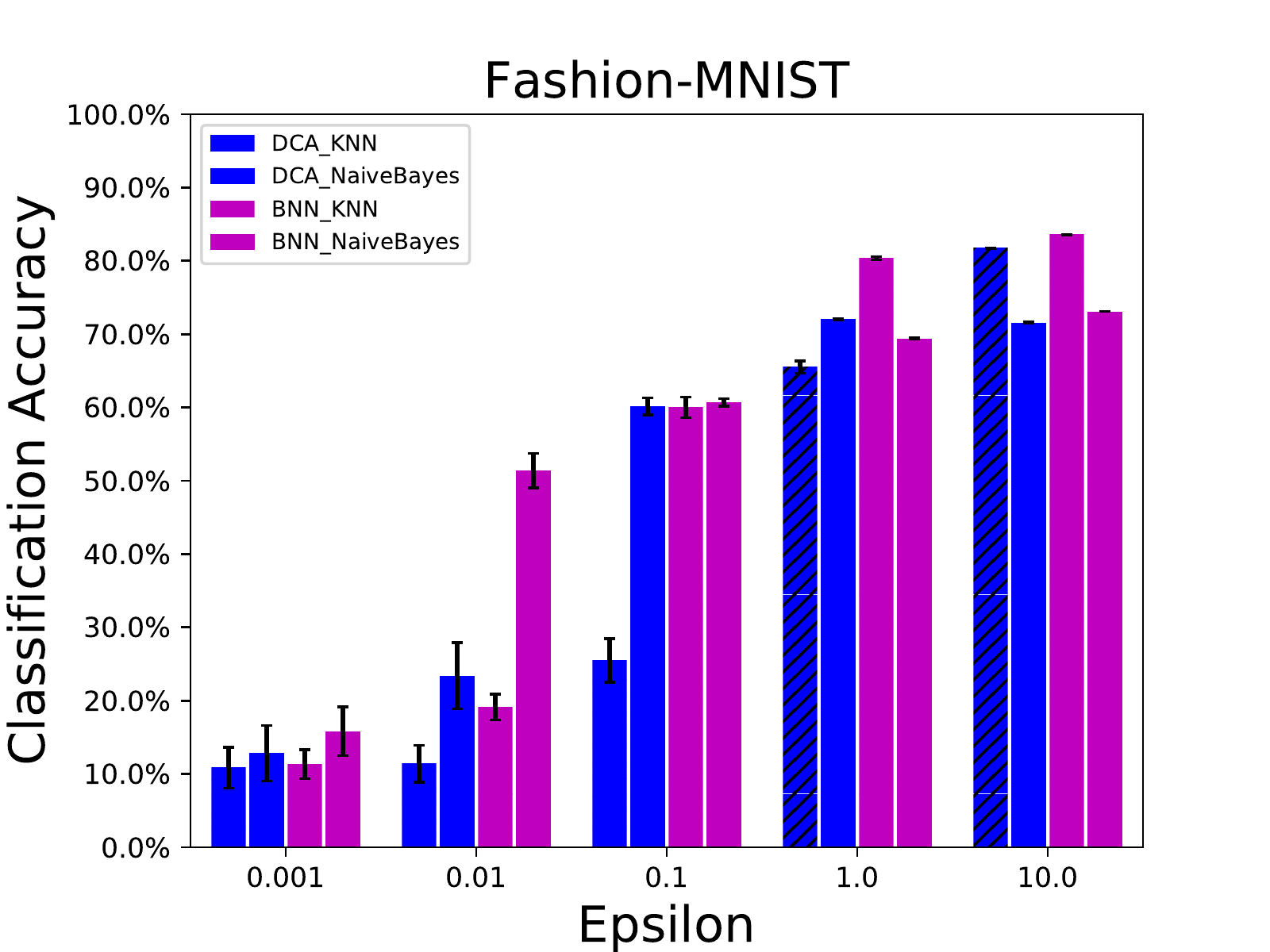}
\caption{}
\end{subfigure}
\begin{subfigure}[t]{.32\textwidth}
\centering
\includegraphics[width=1\textwidth]{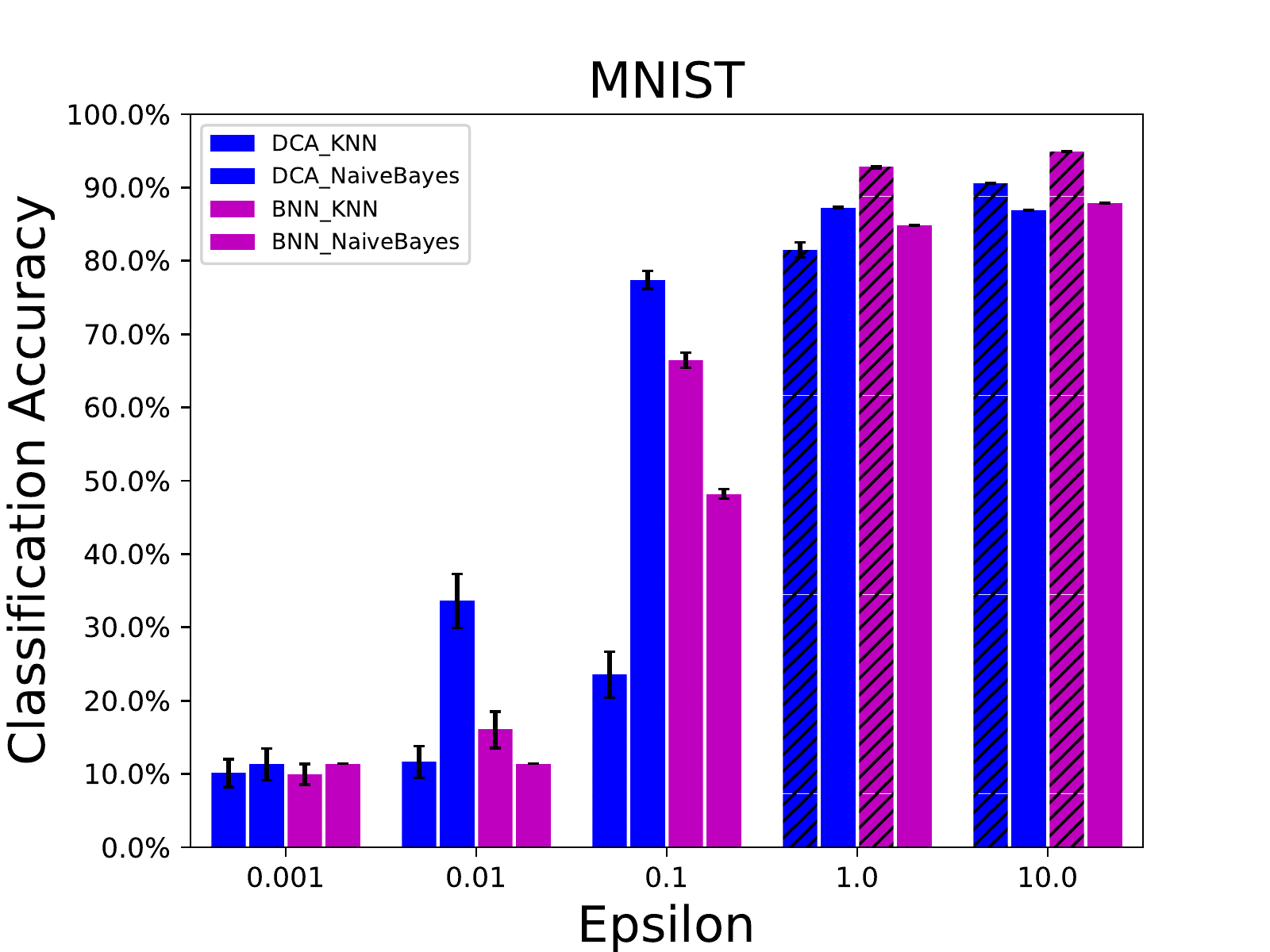}
\caption{}
\end{subfigure}
\begin{subfigure}[t]{.32\textwidth}
\centering
\includegraphics[width=1\textwidth]{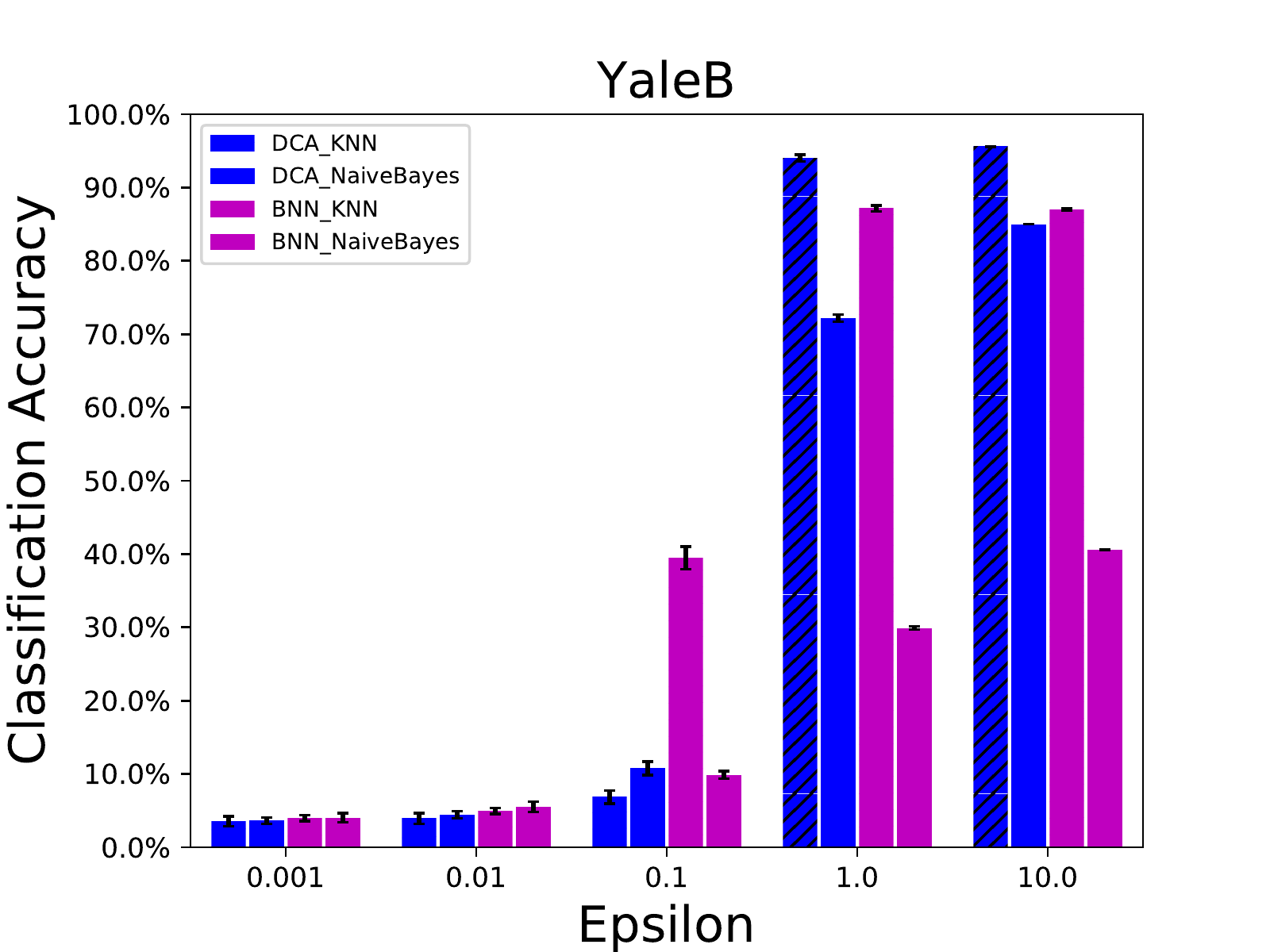}
\caption{}
\end{subfigure}
\end{subfigure}
\medskip
\caption{Comparison of DCAConv and BNN, DCA\_KNN represents result of DCAConv output in KNN algorithm, BNN\_KNN represents result of BNN output in KNN algorithm, output domain $d$=2, number of neighbors in KNN: 100. DCAConv hyperparameters, filter size: 7$\times$7, filter stride size: 1$\times$1, $L_1$: 5, $L_2$: 1, pooling window size: 2$\times$2, pooling stride size: 1$\times$1.}
\label{fig:dca_vs_bnn}
\end{figure*}
We evaluate DCAConv under an extreme case, where the output is set to binary, and compare the utility of the representation to a variant of CNN. As we mention early, the reason that it is hard to apply randomized response to CNN is that the data flow is continuous. However, there are studies to replace the weights and activation with discrete values for the purpose of improving the power-efficiency. Binarized Neural Network (BNN)\cite{courbariaux2016binarized} is one of such research work, which modifies the convolution layer and activation function, to transform the data flow from float number to $\{+1,-1\}$. For the implementation of BNN, we take the suggested values for parameters like the number of hidden units in the fully connected layer, but with the same convolution layer setting as DCAConv, like the same number of filters. Once the training of BNN is finished, we take the activated binarized output before the fully connected layer and perturb it using randomized response. The classification result is shown in Fig~\ref{fig:dca_vs_bnn}, where the vertical axis specifies the classification accuracy. With the binary representation, the probability of preserving the true proximity quickly dominates as $\epsilon \geq 1.1$, and the performance of the perturbed data almost reaches the ground truth. Similarly, Tab.~\ref{tab:dcaconv_d_2} provides the Naive Bayes and KNN accuracy of DCAConv when $\epsilon \geq 0.1$, where the domain size $d=2$, it can be seen that the classification accuracy of KNN almost matches the ground truth regarding all 3 datasets when $\epsilon \geq 1.5$. However, unlike KNN, there is not a clear threshold of $\epsilon$ for Naive Bayes classifier to provide a utility guarantee.

\section{Related Work}

Image privacy has been studied widely. Privacy-aware image classification has been addressed in \cite{zerr2012privacy} and \cite{spyromitros2016personalized}, where they are interested in determining if an image has privacy-related content or not. By exploring the ``public/private'' tags provided by the end users, they trained a privacy classifier to help the end users make decisions in the context of image sharing. A privacy-preserving image feature extraction system is proposed in \cite{qin2014towards}, namely SecSIFT, which allows the user to delegate the SIFT feature extraction to the cloud computing platform. In the design, the image owners encrypt the image, then upload the encrypted data to the \textit{Cloud}. And the computation is distributed to multiple independent \textit{Cloud} entities, who are not colluding with each other. Once the feature extraction is finished, the encrypted feature descriptors are returned to the image owners. The authors argue that both the plaintext data and the locations of feature points of the image are not leaked to the \textit{Cloud}. To the best of our knowledge, the most similar work to this paper is \cite{erkin2009privacy}, a privacy-enhanced face recognition system. They consider a two party problem, the server owns a database of face images, and the client who holds a facial image. The protocol they propose allows the client efficiently hiding both the biometrics and the result from the server that performs the matching operation by using secure multiparty computation technique. The difference from our work is that it is a centralized setting where the data has already been collected in the server. However, the technique (e.g, without adding noise for differential privacy) provides inadequate privacy protection.

LDP is firstly proposed by \cite{duchi2013local}, in where the minimax bounds for learning problem is derived under the local privacy definition. Kairouz et al. \cite{kairouz2014extremal} study the trade-off between the utility and LDP in terms of family of extremal mechanisms. Although two simple extremal mechanisms, the binary and randomized response mechanisms, are proposed and shown the theoretical guarantee of the utility, it is still hard to adopt the methods in real world machine learning problems. Wang et al. \cite{wang2017locally} propose a framework that generalizes several LDP protocols in the literature and analyzes each one regarding frequency estimation. For the real world application, RAPPOR\cite{erlingsson2014rappor} is the first that practically adopts LDP to collect the statistics from the end users, which is designed by Google. Apple and Microsoft also propose their own tools to use LDP in collecting usage, typing history and telemetry data\cite{ding2017collecting}. 

Recently, an iterative method\cite{nguyen2016collecting} that satisfies $\epsilon$-LDP is proposed for empirical risk minimization, where each data owner locally computes and perturbs the stochastic gradient descent (SGD) of the parameters, then shares the noisy version of SGD to the aggregator. In the paper, the authors demonstrate the computation with three models, the linear regression, logistic regression and SVM. Another client-server machine learning framework is proposed to learn the Generalized Linear models\cite{pihur2018differentially}, where the server simultaneous delivers the models to the distributed clients and asks for the parameter update. More specifically, the server maintains $k$ versions of the machine learning model, and randomly selects one version to update the parameters and replaces the another version. To enforce the $\epsilon$-LDP, the client uses Laplacian mechanism to perturb the parameters and returns the model to the server. Unlike both works, we consider a ``non-interactive'' scenario where the data owners only perturb the input once and then share the noisy input with the data user. Comparing to the interactive fashion, the non-interactive way requires less communication cost and brings a great scalability. Overall, most current studies regarding LDP either focus on the theoretical interest of the perturbation methods or on simple statistical collection, like mean and frequency estimation. There are few studies regarding machine learning in a local setting.
\section{Conclusion}
In this paper, we study a privacy-preserving image classification problem, where the distributed data owners hold a set of images; the untrustworthy data user would like to fit a classifier regarding the images held by the owners. To protect the privacy, we propose to use randomized response to perturb the image representation locally, which satisfies $\epsilon$-Local Differential Privacy. We address two questions in this paper, firstly, we analyze the utility and privacy trade-off regarding the domain size $d$ and parameter $\epsilon$ for Naive Bayes and KNN classifier. For Naive Bayes classifier, the utility is bounded by the frequency estimation of the individual feature; for KNN, we show that the probability of retaining the true proximity between the testing sample and the perturbed training sample is bigger than $\frac{1}{2}$ as $\epsilon \geq ln(d+1)$; secondly, to balance the trade-off between the utility and privacy, we propose a supervised image feature extractor, namely DCAConv, which produces a representation of the image with a scalable domain size. The experiments  over three benchmark image datasets evaluates the classification accuracy of the perturbed image representations with respect to Naive Bayes and KNN algorithm, which confirms our analysis regarding the domain size and $\epsilon$ and shows that the effectiveness of DCAConv under various privacy constraints.

\bibliographystyle{aaai} \bibliography{./manuscript/citation}
\end{document}